\documentclass[11pt,letterpaper]{article}
\usepackage{emnlp2017}
\usepackage{latexsym}
\usepackage{CJK}
\usepackage{graphicx}
\usepackage{multirow}
\usepackage{url}
\usepackage{makeidx}  
\begin{document}
\pagestyle{headings}  

\title{Sentence Correction Based on Large-scale Language Modelling}
\emnlpfinalcopy
\author{Ji Wen \\School of EECS, Peking University\\      {wenjics}@pku.edu.cn }

\maketitle              

\begin{abstract}
With the further development of Internet, more and more data is stored in the form of text. There are some omission of text during their generation and transmission. The paper aims to establish a language model based on the large-scale corpus to complete the restoration of missing word. In this paper, we introduce a novel measurement to find the missing words, and a way of establishing a comprehensive candidate lexicon to insert the correct choice of words. The paper also introduces some effective optimization methods, which largely improve the efficiency of the text restoration and shorten the time of dealing with 1000 sentences into 3.6 seconds. 

\end{abstract}
\section{Introduction}

With the progress of information technology and the further popularization of the Internet, the digitized text has become the main source of information. At the same time, the texts in the traditional form are being digitized. However, whether being scanned in the ways such as OCR, or being manually inputted, the occurrence of errors cannot be avoided. And now in the field of natural language processing, as a research project and the results achieved is few. It is difficult to find closely related literature even on the Internet. With the further explosion of data, the restoration of the valuable part of texts is believed to receive more attention. The information after the repair, can also provide more reliable experimental data for other experiments.

In this article, the completion of the sentence for missing words is considered. The size of training data is up to 3.86GB. Based on this large-scale corpus, a statistical model will be established to accomplish the sentence correction for missing words in the test data. In the testing process, the correct inserting position of the word need to be found, as well as the correct word to complete the insertion. For example, for sentences:

\emph{He added that people should not mess with mother nature , and let sharks be.}

Firstly, need to find that the location of the insertion is between the ``let" and ``sharks'', and then from all the candidate words, select the correct word ``the" to insert.
%
%
%

Therefore, the testing process will be two steps. The first step is to find the correct position to insert the word The second step is to find suitable words to insert. In order to elaborate the methods used in the experiment, the organization of the chapters is as follows: the second chapter introduces the related work; the third chapter introduces the method proposed in this paper; the fourth chapter gives the experimental results; Five chapters introduce the optimization used in the experiment; the final will give a summary of the full text.

\section{Proposed Method}

The main part of the experiment has two steps. One is to find the position to insert the word, the second is to choosing the appropriate word to insert.
First of all, for finding inserting location, the most intuitive method is based on the relevance of the adjacent words. The probability distribution provided by the n-gram language model is not sufficient as the determination of the insertion position. For example, suppose $w_{1}$, $w_{2}$ as a common phrase, where $w_{1}$ is the high frequency word, $w_{2}$ is the low frequency word, and the $w_{2}$ use scenario is almost entirely used as the phrase $w_{1}w_{2}$. The result is that in the binary language model, the probability $P(w_{2}|w_{1})$ will be very small, if the sentence as a basis for the insertion, then it is likely to separate the phrase. So we need to distinguish it from the conventional multivariate language model the way.

Secondly, it is also difficult to choosing the appropriate word, since there is not a list of candidate words. All words can be candidate words. It needs to reduce the size of the list of candidate words as much as possible, but still with suitable words included. In the meantime, the words that are truly missing in the sentence can be broadly divided into two parts. The first part is with the actual meaning, which will affect the sentence ideology, such as ``sharks'' and ``husband''. The other part is not with too much meaning, which acts as part of the sentence structure, such as ``the''.
\subsection{Constructing Static Candidate Thesaurus}
As mentioned before, the missing words in the test are two kinds. The second is mainly for the existence of the syntactic structure of the sentence. These are often high frequency words, such as ``the''. It should be emphasized that the words inserted in the experiment may also be punctuation. High frequency punctuation such as comma and quotation marks are usually high frequency words.

\subsection{Constructing the ``Together'' Table and ``Separate'' Table}
In order to find the correct inserting position, it needs to analyze the relevance of the words in each sentence of the test data. Here set up the ``together" table and the ``separation'' table as a basis for judging.

Firstly, for the test sentence, get its adjacent phrase, $w_{1}w_{2},...,w_{m}$. For each binary word combination $w_{i}w_{i+1}$, from the training data, calculate the frequency of $w_{i}$ and $w_{i+1}$ adjacent, denoted together ($w_{i}w_{i+1}$). Then calculate the frequency of $w_{i}$ and $w_{i+1}$ when there is just one word between them, recording it as separate ($w_{i}w_{i+1}$).

\subsection{Choosing Inserting Position}
Choosing inserting position depends on the ``together'' table and the ``separate" table. In this experiment, it uses the division of them to determine the ``separation probability'' of the binary phrase $w_{i}w_{i+1}$.The higher the ratio of separate ($w_{i}w_{i+1}$) to together ($w_{i}w_{i+1}$), the greater the likelihood that it should be the insertion position.

$HYPER\_V$ is set as hyper parameter to determine whether the value of separate ($w_{i}w_{i+1}$) together ($w_{i}w_{i+1}$) is high enough to cause an insertion. 

\subsection{Constructing Dynamic Candidate Thesaurus}
It constructs a dynamic candidate thesaurus based on the ``separation" table. When the insertion position is in the middle of $w_{i}w_{i+1}$, it will choose words that have great relationships with $w_{i}w_{i+1}$ to form a part of the thesaurus. The dynamic candidate thesaurus can actually be seen as a mapping of $w_{i}w_{i+1}$ to a set of words appear between $w_{i}$ and $w_{i+1}$ in the training data.
\subsection{Choosing the Inserting Word}

For each possible insertion position, try the static thesaurus and the corresponding dynamic thesaurus. Then calculate the probability of the sentence based on the previously obtained trigram. Finally, select the maximum probability for the final result.


\section{Experiments and Results}

The machine used in the experiments is with i5-3740 CPU and 16 GB memory.

The training data is from a standard corpus presented by Cornell University in 2013 [12]. 
The corpus size is about 3.86GB, which contains the number of words close to one billion.

There were 306,681 sentences in the test data, and each sentence was removed with a single word. The removed word will not be at the beginning of the sentence, nor will it be at the end of the sentence (in this test set, the suffix will always be a full stop). In addition, the position of the removed word is randomly selected.
The results of the program will be compared with the correct answer, the evaluation criteria is the average Levenshtein distance between them.

\begin{figure}
  \centerline{\includegraphics[width=9cm]{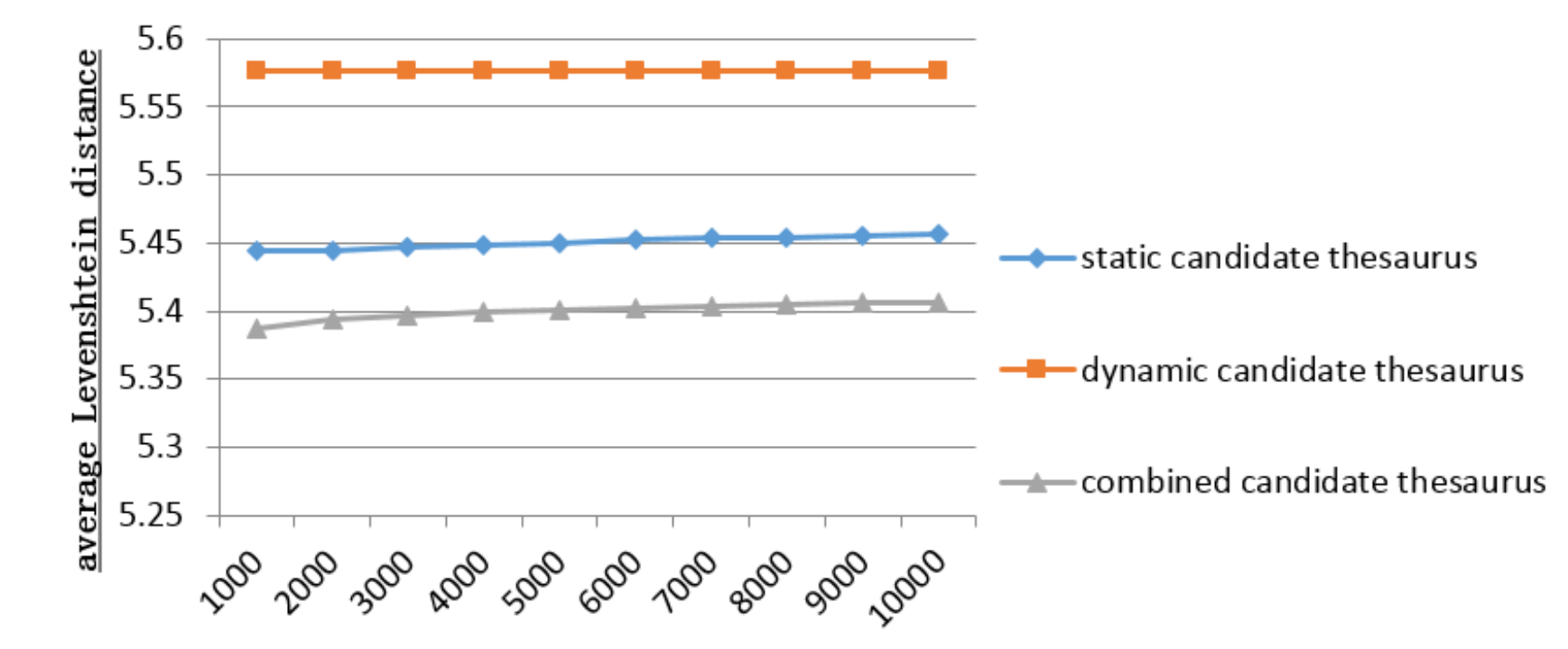}}
  \caption{The experiment results of three candidate thesauruses. }
\label{asynchornous}
\end{figure}
\subsection{Locating the Inserting Position}
In the experimental design section, the search for the insertion position has been described in detail. In the actual process, determining the $HYPER\_V$ value is an exploratory process. If $HYPER\_V$ is selected too small, it will lose the restriction on separate($w_{i}w_{i+1}$)/together($w_{i}w_{i+1}$), resulting in an insert action where the two adjacent words should not be separated. Conversely, if $HYPER\_V$ is chosen too large, it will exclude too many locations where the insertion is more likely, resulting in the sentence giving up the insertion action. So the search process for $HYPER\_V$ is also a challenge. By comparing multiple experiments, $HYPER\_V = 27$ was chosen in the actual experiment.

\subsection{Choosing the Inserting Word}
Compared to finding inserting position, choosing the inserting word is much more different. Firstly, the range of candidates for the insertion position is limited and definite, but the range of candidates for the inserted words covers all the words no matter if they appear or not in the training data and the test data. Secondly, the error in choosing word error causes less damage than the error in finding inserting position. Based on these two points, we expect to be able to use enough lexicon to get enough good results.

Only using dynamic candidate thesaurus lead to the worst results. The reason may be over-fitting. On the contrary, the results of only using static candidate thesaurus are not bad. And after combing of these two thesauruses, the effect of the upgrade is significant. Regardless of the size of the static candidate thesaurus, the overall effect is better than their separate use.

\subsection{Experimental Optimization}
In the course of the experiment, it is found that the occupancy rate of memory has been relatively high, but the CPU utilization has been only about 30\% hovering.

 Thus parallelization is considered. By packing multiple sentences into a task as a thread, the results will be a one-time output to the file. Since the insertion of each sentence does not affect each other, the shared resource only involves reading. There is no need to worry about the multi-threaded conflict, but the synchronization still needs some time to consume.

For the final test process, the serial program averages 7.8s per 1000 words, and the average of the parallel program is 3.6s per 1000.

\section{Related Work}

\subsection{Language Model}
The language model plays an important role in natural language processing. The statistical language model depend on the context environment. In order to simplify the context as much as possible, there has been a decision tree language model \citep{DBLP:journals/tsp/BahlBSM89}. In order to solve the problem of data fragmentation, the maximum entropy model 
has also been proposed. At present, this paper uses the n-gram language model.
At the same time, since the size of the corpus is always limited, smoothing algorithms must be used. After a long period of research, there are many smoothing algorithms: additive smoothing algorithm, Gould-Turing estimation method \citep{good1953population}, Katz smoothing method \citep{DBLP:journals/tsp/Katz87}, Jelinek-Mercer smoothing method \citep{jelinek1980interpolated} and so on. In this experiment, Katz smoothing method is used. In the future, it would be interesting to explore the usage of neural language models or sequence to sequence neural models  \citep{DBLP:conf/acl/MaSXWLS17,shumingma} to improve the performance of language models in this task.

\subsection{Sentence Correction}
Sentence corrections difficult to find a reliable paper to refer to. The reason is that the language model is now mostly statistical language model, which retain a certain degree of tolerance for a few errors in the data. On the other side, sentence correction can be regarded as a "translation" process. Based on statistical methods, machine translation has a lot of research results \citep{DBLP:conf/acl/1997} \citep{DBLP:journals/coling/BrownPPM94}, which also contains the phrase as the basic unit of translation \citep{DBLP:conf/naacl/KoehnOM03}. These translations give the experiment some of the angles that can be considered. For the misspelled corrections, there are methods based on the noise channel \citep{DBLP:conf/acl/BrillM00}, there are other methods based on statistics \citep{DBLP:journals/ipm/AngellFW83} and text \citep{DBLP:journals/ipm/MaysDM91}. There are also many similarities with the task of query spelling correction \citep{DBLP:conf/interspeech/ChelbaMSGBKR14}. Learning phrase-based spelling error model \citep{DBLP:journals/corr/Magerman94}  may help the task of sentence correction as well.  The use of clickthrough data was explored for query spelling correction \citep{DBLP:conf/coling/GaoLMQS10,DBLP:conf/acl/SunGMQ10}. Query spelling correction using multi-task learning was also proposed for optimizing the performance \citep{DBLP:conf/www/SunSL12,DBLP:conf/cikm/SunSL12}. But the method proposed in this paper is different from them, the following will be described in detail.

\section{Conclusion}

For the insertion of words in sentence correction, this paper has done some experiments on the process of finding the inserting position and the process of finding the inserting word. In general, there are three main achievements in this paper. One is finding a lightweight and convenient method to determine whether the binary combination should be ``separated''. The second is a combination of static candidate thesaurus and dynamic candidate thesaurus, in the search for the inserting word. The third is the attempt of parallelization, making the time of test process greatly shortened.
The research results of this paper have made some attempts for word insertion in sentence correction, and the methods and results used have value for the follow-up research.

\bibliography{emnlp2017}

\bibliographystyle{emnlp_natbib}

\end{document}